\documentclass[conference]{IEEEtran}
\IEEEoverridecommandlockouts
\usepackage{cite}
\usepackage{amsmath,amssymb,amsfonts}
\usepackage{graphicx}
\usepackage{textcomp}
\usepackage{xcolor}
\usepackage{url}
\usepackage{tikz}
\usepackage{eurosym}
\graphicspath{{figures/}}

\begin{document}

\makeatletter
\newcommand{\linebreakand}{%
  \end{@IEEEauthorhalign}
  \hfill\mbox{}\par
  \mbox{}\hfill\begin{@IEEEauthorhalign}
}
\makeatother

\title{Data-Driven Configuration of Protective Separation Zones in Collaborative
Robot Cells from Operator Occupancy}

\newif\ifblindreview
\blindreviewfalse

\ifblindreview
\author{\IEEEauthorblockN{Anonymous Authors}
\IEEEauthorblockA{Affiliation withheld for double-blind review}}
\else
\author{\IEEEauthorblockN{Aulon Bajrami}
\IEEEauthorblockA{\textit{Fraunhofer Institute for Manufacturing}\\
\textit{Engineering and Automation IPA}\\
Stuttgart, Germany\\
aulon.bajrami@ipa.fraunhofer.de}
\and
\IEEEauthorblockN{Sharath Nataraj}
\IEEEauthorblockA{\textit{Fraunhofer Institute for Manufacturing}\\
\textit{Engineering and Automation IPA}\\
Stuttgart, Germany\\
sharath.nataraj@ipa.fraunhofer.de}
\and
\IEEEauthorblockN{Mrunal Sompura}
\IEEEauthorblockA{\textit{Fraunhofer Institute for Manufacturing}\\
\textit{Engineering and Automation IPA}\\
Stuttgart, Germany\\
mrunal.sompura@ipa.fraunhofer.de}
\linebreakand
\IEEEauthorblockN{Theo Jacobs}
\IEEEauthorblockA{\textit{Fraunhofer Institute for Manufacturing}\\
\textit{Engineering and Automation IPA}\\
Stuttgart, Germany\\
theo.jacobs@ipa.fraunhofer.de}
\and
\IEEEauthorblockN{Werner Kraus}
\IEEEauthorblockA{\textit{Fraunhofer Institute for Manufacturing}\\
\textit{Engineering and Automation IPA}\\
Stuttgart, Germany\\
werner.kraus@ipa.fraunhofer.de}}
\fi

\maketitle
\begin{tikzpicture}[remember picture,overlay]
\node[anchor=north,yshift=-0.28in,text width=\textwidth,align=justify,font=\scriptsize]
  at (current page.north) {Accepted for publication at the 2026 IEEE Stuttgart Conference on Automotive Production (SCAP).
\copyright~2026 IEEE. Personal use of this material is permitted. Permission from IEEE must be obtained for all other uses, in any current or future media, including reprinting/republishing this material for advertising or promotional purposes, creating new collective works, for resale or redistribution to servers or lists, or reuse of any copyrighted component of this work in other works.};
\end{tikzpicture}

\begin{abstract}
The protective separation zone of a collaborative robot cell is commonly
dimensioned for a worst case that disregards where the operator is actually
present during the task. This paper presents a camera-based, data-driven method
that configures this zone from observed operator occupancy. Operator positions
accumulated over several production shifts are aggregated into an occupancy
heatmap, from which a protective field is derived that trades off the achievable
robot velocity against the required safety distance, following the speed and
separation monitoring principle of ISO/TS~15066. The resulting configuration is
advisory, that is, reviewed within the risk assessment rather than applied
autonomously to the robot controller. The method, which builds on a pose-based
perception pipeline with activity recognition, was evaluated in two operational
cells at two industrial partners, where the protective field could be reduced by approximately 17\% and
a productivity increase of 9\% was measured. However, adoption proved to depend
on more than the demonstrated gain alone.
\end{abstract}


\section{Introduction}

Collaborative robots promise to combine the repeatability of automation with the
flexibility of the human operator, and they have consequently attracted
considerable attention as a means of automating tasks that resist full
automation, particularly in small and medium-sized enterprises
\cite{villani2018survey,matheson2019hrc,vysocky2016hrc}. Realising this promise
in production, however, requires the shared workspace to be kept safe without
sacrificing the very productivity that motivates the deployment
\cite{wang2019symbiotic,gualtieri2021emerging}. The present work is concerned
with this tension in cells that implement speed and separation monitoring in the
sense of ISO/TS~15066 \cite{isots15066_2016}.

The contribution of this work is twofold. First, a data-driven method for
configuring the protective separation zone of a collaborative cell is presented:
operator positions observed over time are aggregated into an occupancy heatmap,
from which an optimised size of the protective field is derived that trades off
the achievable robot velocity against the required safety distance, following the
speed and separation monitoring principle of ISO/TS~15066. The result is
advisory, that is, it is provided as a proposal reviewed within the risk
assessment before adoption rather than applied autonomously to the robot
controller. Second, the perception pipeline from which the occupancy is obtained
is described, comprising pose-based person tracking and an activity recognition
step that distinguishes interaction with the cell from unrelated traffic. Unlike prior fluency and safety approaches that
are validated in simulation or in the laboratory under idealised perception
(e.g., \cite{scalera2022}), the proposed approach was developed and evaluated
together with industrial partners in a real production setting, where occupancy
estimation and activity recognition must contend with the occlusions, layouts
and operator behaviour encountered on the shop floor.

\section{Related Work}

The economic case for collaborative robotics is increasingly constrained by the
effort required to keep shared workspaces safe. In a recent study of German
system integrators, 75\% of the respondents reported
that safety-related standards limit the achievable performance of collaborative
robots, with ISO/TS 15066 named most frequently (45.8\%), and 43.8\% of all
reported implementation problems were found to relate directly to the risk
assessment, which is perceived as time-consuming and largely subjective
\cite{kifz2025mrk}. The same study indicates a clear demand for data-driven
support, as 66\% of the respondents stated that safety-relevant data would help
them during risk assessment. These findings motivate the
present work, which obtains such data from a camera-based observation of the
cell. It should be noted, however, that the same
study also reports reservations regarding privacy and standard conformity, which
a practical deployment has to address.

A substantial body of work addresses safety and fluency in human-robot
collaboration (HRC) from the side of robot control
\cite{zanchettin2016safety,villani2018survey}. A common strategy is the
online adaptation of safety zones around the manipulator in order to reduce
unnecessary stops while preserving separation, whether by scaling the commanded
speed along the planned path \cite{zanchettin2013path}, by planning
human-aware motions that keep the operator at a distance
\cite{lasota2015analyzing}, or by switching between graded interaction regions
around the robot \cite{bdiwi2017new}; the fluency of the resulting interaction
has itself become a subject of study \cite{hoffman2019evaluating}. Scalera et
al. \cite{scalera2022}
propose a supervisory controller that scales dynamic safety zones online and
plans minimum-time stop trajectories from the robot dynamics and its torque
limits, reporting reductions in total task time of up to 26.5\% with respect to
a comparable dynamic method and 60.8\% with respect to a static implementation
of ISO/TS 15066. The separation-distance computation underlying such fields is
specified by ISO/TS~15066 and refined in the 2025 revision of the ISO~10218
series \cite{isots15066_2016,iso10218_2_2025,iso10218_1_2011}, and its practical
realisation in collaborative workcells has been studied in detail
\cite{marvel2017ssm,byner2019dynamicssm}, including its combination with power
and force limiting \cite{lucci2020combining}. Such approaches, however, operate
inside the robot controller
and presuppose that the human pose is already available from a fixed, suitably
placed sensor, and their validation is typically carried out in simulation or in
a single-arm laboratory setup using nominal trajectories rather than real
production tasks \cite{scalera2022}. In contrast, comparatively little attention
has been paid to how the protective separation zone itself is dimensioned in
practice, which is commonly fixed to a worst-case configuration that disregards
where the operator is actually present during the task. Configuring this zone
from operator occupancy observed on the shop floor, and recognising human
activity reliably under realistic occlusions and layouts, has received little
attention. This contribution addresses that gap.

On the perception side, the ingredients required to observe the operator are by
now mature. Multi-person 2D pose can be recovered in real time
\cite{cao2021openpose}, persons can be detected and tracked across frames by
online multi-object trackers \cite{wojke2017deepsort,zhang2022bytetrack}, and
their activity can be classified from video with spatio-temporal convolutional
networks \cite{carreira2017i3d,feichtenhofer2019slowfast} or from the recovered
skeleton with graph convolutional networks \cite{yan2018stgcn}, as surveyed in
\cite{gu2021harsurvey}. Camera- and depth-based sensing has also been applied
directly to the monitoring of safety zones in collaborative cells
\cite{slovak2021vision}. The present work builds on these components but is
concerned less with the recognition itself than with turning the resulting
occupancy statistics into a configuration decision for the protective field.

\section{Methodology}

The proposed system is realised as an edge camera unit, which is mounted in an
existing collaborative cell and observes the shared workspace. It provides two outputs:
(i) a recognised activity label for the operator, and (ii) an estimate of the
number and the approximate position of the persons present in the vicinity of
the cell. These outputs are accumulated over an observation period of several
production shifts, from which an occupancy model of the cell is derived. On the
basis of this model, a modified configuration of the protective fields is
proposed, which is subsequently reviewed by the safety engineer and commissioned
in the ordinary manner. The unit therefore neither participates in the motion
control of the manipulator nor forms part of the safety-rated signal path, which
is what allows it to be retrofitted to an existing cell. An overview of the
approach is given in Fig.~\ref{fig:concept}.

\begin{figure*}[tb]
\centering
\begin{tikzpicture}
\node[anchor=south west, inner sep=0] (img)
  {\includegraphics[width=\textwidth]{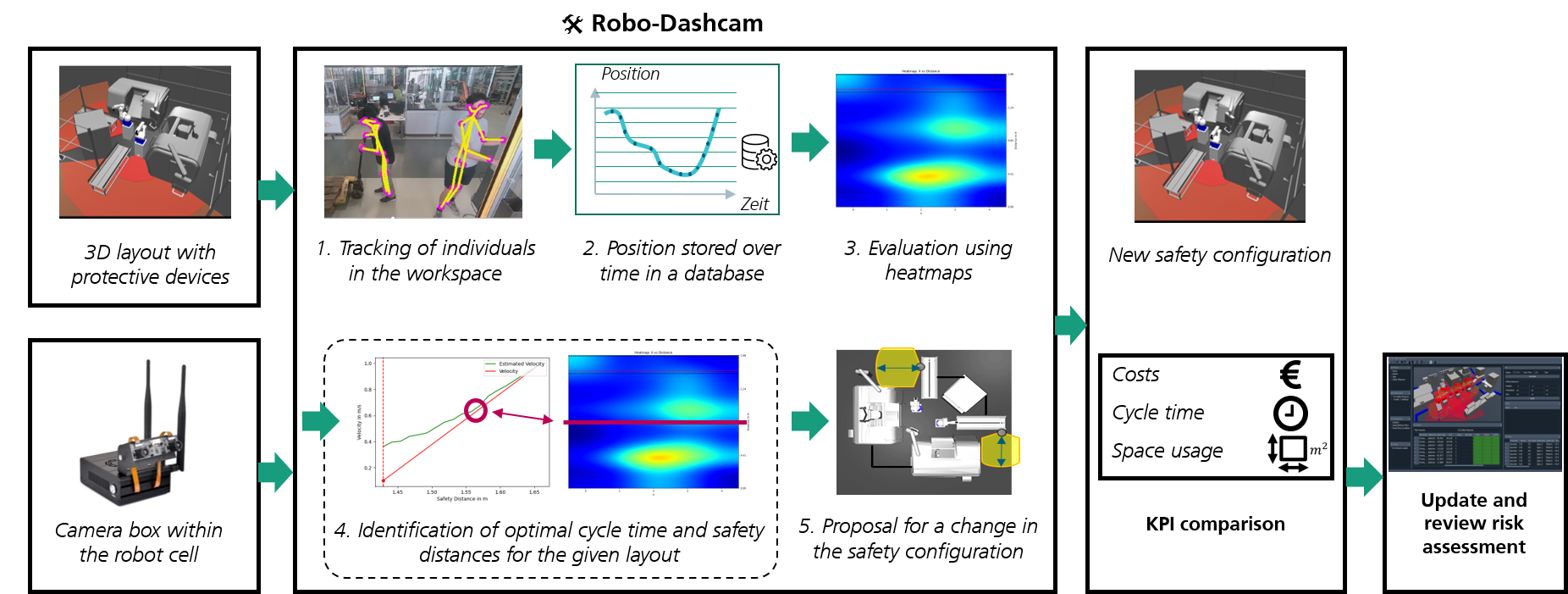}};
\begin{scope}[x={(img.south east)}, y={(img.north west)}]
\ifblindreview\fill[white] (0.350,0.934) rectangle (0.495,1.0);\fi
\end{scope}
\end{tikzpicture}
\caption{Overview of the proposed approach. Starting from the layout of the cell
and the camera box installed within it, the persons present in the workspace are
tracked (1) and their positions are accumulated over time (2), from which
occupancy heatmaps are obtained (3). The admissible speed and the associated
safety distance are derived for the given layout (4), and a modified safety
configuration is proposed (5) and compared against the previous one in terms of
cost, cycle time and space usage.}
\label{fig:concept}
\end{figure*}

\subsection{Perception Layer}
The camera unit is equipped with an RGB-D camera (Intel RealSense, 640~$\times$~480 px,
30 fps), which is mounted such that the area in front of the cell, including the
protective and warning fields, lies within its field of view. The pose of the
camera relative to the cell is determined once during installation, so that all
positions can be expressed in the coordinate frame of the cell layout. A
related setup is described in \cite{bajrami2024multi}, where the depth data of
several RGB-D cameras were fused into an occupancy map of a collaborative cell
and the robot speed was scaled by the distance between the operator and the
robot. In contrast to that work, which evaluates the occupancy
instantaneously with a depth-based skeleton tracker, the present pipeline
estimates the pose with a learned model and accumulates the positions over time.
The perception pipeline comprises four stages. Firstly, the 2D pose of every person
in the image is estimated with Lightweight OpenPose \cite{osokin2019lightweight},
a MobileNet-based variant of OpenPose \cite{cao2021openpose}, which is used with
its publicly released weights without retraining and yields 18 keypoints per
person in the COCO layout. Secondly, the poses are associated across frames by
an Object Keypoint Similarity (OKS) measure, so that every person retains an
identity over time, and the keypoints are smoothed with a One Euro filter
\cite{casiez2012oneeuro} in order to suppress jitter without introducing
noticeable lag. Thirdly, every valid keypoint is lifted to 3D by reading its
depth from the aligned depth image and deprojecting it with the pinhole model of
the camera. Where the depth is invalid, for instance at edges or on reflective
surfaces, a valid value from the neighbourhood of the pixel is used. Finally,
the skeleton is projected onto the floor plane of the cell, which provides the
position of the person, while the number of tracked identities within the
observed volume provides $n(t)$. On the basis of the tracked persons, the
activity of each person is recognised and labelled as either directed at the
cell or as unrelated traffic passing through the observed area. This label is
used in Sect.~\ref{sec:selection} to tell intended interaction apart from the
traffic that causes avoidable protective stops.

\subsection{Speed and Separation Monitoring}
The cells considered in this work implement speed and separation monitoring in
the sense of ISO/TS 15066, using a horizontally mounted laser scanner that
supervises a protective field and a surrounding warning field. This behaviour is
already installed in the cell and is modelled, rather than implemented, by the
proposed method. Let $d$ denote the smallest distance measured between the
operator and the robot. The speed of the manipulator is scaled by a factor
$s(d)$, which is obtained from a linear ramp between the two fields:

\begin{equation}\label{eq:ramp}
s(d) =
\begin{cases}
0, & d \le d_{0},\\[4pt]
\dfrac{d - d_{0}}{d_{1} - d_{0}}, & d_{0} < d < d_{1},\\[6pt]
1, & d \ge d_{1},
\end{cases}
\end{equation}

with $d_{0}$ being the boundary of the protective field and $d_{1}$ the boundary
of the warning field, and the commanded speed being
$v = s(d)\,v_{\mathrm{nom}}$, where $v_{\mathrm{nom}}$ is the nominal speed of
the fastest point of the manipulator. The robot is therefore brought to a
monitored stop when the operator enters the protective field, and travels at
nominal speed once the warning field is clear. More elaborate scaling laws could
be adopted. However, a linear ramp was found to be sufficient in the cells
considered here, and it has the practical advantage of remaining comprehensible
to the safety assessor during acceptance testing.

\begin{figure}[tb]
\centering
\begin{tikzpicture}
\node[anchor=south west, inner sep=0] (img)
  {\includegraphics[width=\columnwidth]{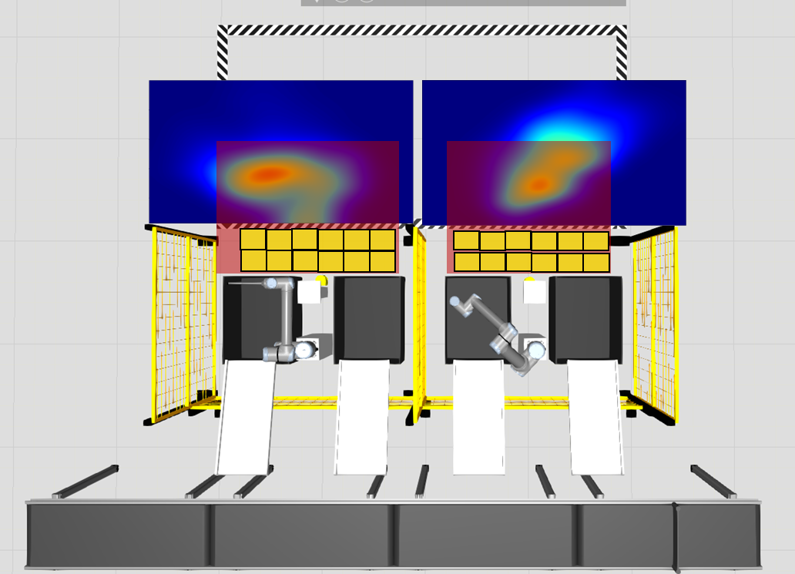}};
\begin{scope}[x={(img.south east)}, y={(img.north west)}, font=\scriptsize]
\tikzset{callout/.style={-latex, line width=0.6pt,
  preaction={draw=white, line width=1.8pt, -}}}
\node[anchor=west] (wf) at (0.02,0.935) {Warning field};
\draw[callout] (wf.east) -- (0.276,0.935);
\node[align=center] (pfl) at (0.09,0.72) {Protective\\field};
\draw[callout] (pfl.east) -- (0.29,0.65);
\node[align=center] (pfr) at (0.925,0.72) {Protective\\field};
\draw[callout] (pfr.west) -- (0.75,0.65);
\end{scope}
\end{tikzpicture}
\caption{Operator occupancy heatmaps for the two collaborative cells considered
in the beta tests, projected into the cell geometry. Warm colours denote a high
relative frequency of operator presence, cold colours denote regions that are
rarely entered. The protective field of each cell is shown in translucent red,
and the warning field is indicated by the hatched boundary.}
\label{fig:heatmap}
\end{figure}

\begin{figure*}[tb]
\centering
\begin{tikzpicture}
\node[anchor=south west, inner sep=0] (img)
  {\includegraphics[width=\textwidth]{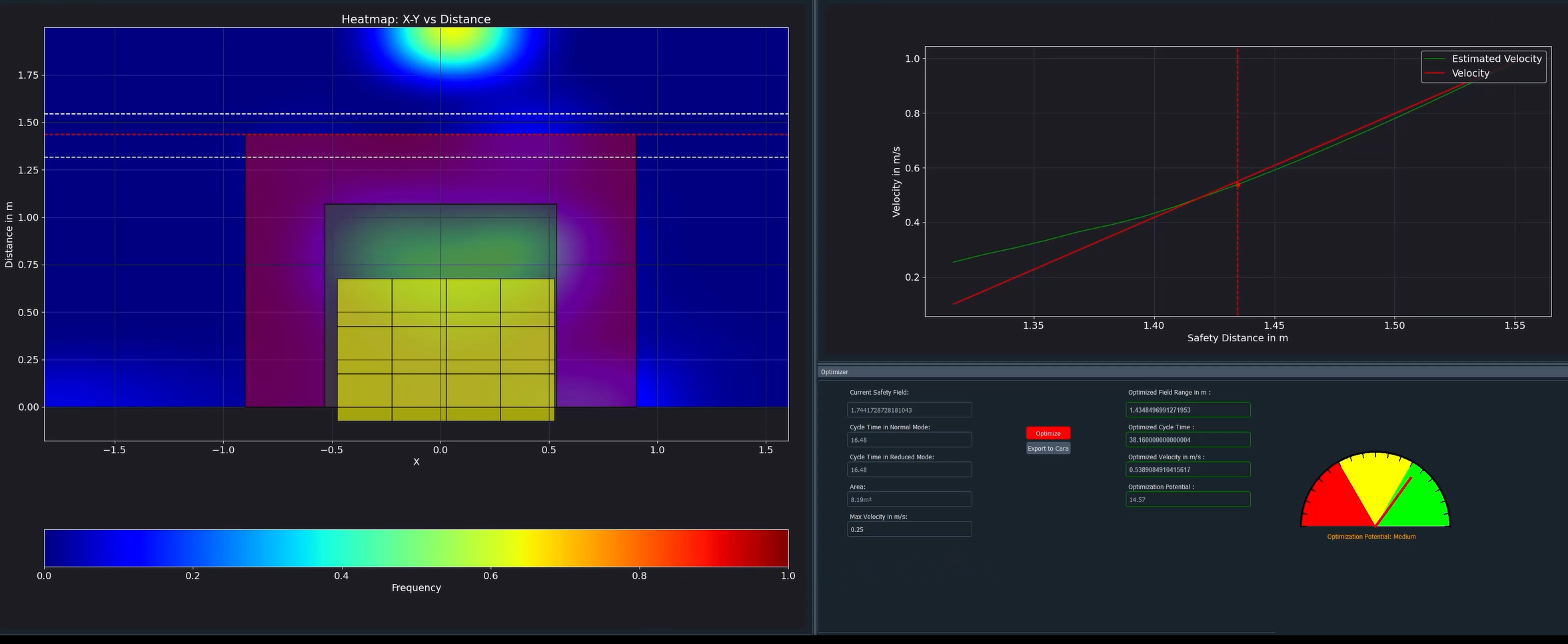}};
\begin{scope}[x={(img.south east)}, y={(img.north west)}]
\definecolor{uipanel}{RGB}{25,35,44}
\fill[uipanel] (0.652,0.290) rectangle (0.685,0.316);
\end{scope}
\end{tikzpicture}
\caption{Safety zone optimisation for one beta-test cell. Left: current and
proposed protective field boundary overlaid on the operator occupancy heatmap.
Right: permissible robot velocity as a function of the safety distance, nominal
and estimated. The optimiser plots the heatmap over lateral position and
distance to the cell, so its representation differs from Fig.~\ref{fig:heatmap}.}
\label{fig:optimizer}
\end{figure*}

\subsection{Field Geometry}
The radius $R$ of the protective field cannot be chosen freely, since it has to
cover the distance travelled by the robot and by the operator during the
reaction time of the safety chain and during the subsequent braking. Under the
assumption of a constant admissible deceleration $a_{\mathrm{max}}$, the stop
time follows as $t_{\mathrm{s}} = v / a_{\mathrm{max}}$, and the field has to
satisfy:

\begin{equation}\label{eq:field}
R \;\ge\; v\,t_{\mathrm{r}}
\;+\; v_{\mathrm{h}}\left(t_{\mathrm{r}} + \frac{v}{a_{\mathrm{max}}}\right)
\;+\; \frac{v^{2}}{2\,a_{\mathrm{max}}}
\;+\; C
\end{equation}

where $t_{\mathrm{r}}$ is the reaction time of the safety chain (scanner
response, PLC cycle, and brake engagement), $v_{\mathrm{h}} = 1.6$ m/s is the
worst-case walking speed of the operator specified in ISO~13855
\cite{iso13855_2010}, and $C$
lumps the measurement tolerance of the scanner together with the portion of the
operator that may intrude into the field before being detected.

Equation~\eqref{eq:field} is quadratic in $v$, and it gives rise to a trade-off which, in the
present experience, is more consequential on the shop floor than the residual
conservatism of the field radius itself. A large field permits a high speed, but
it is intruded frequently, since it extends into regions through which operators
pass without any intention of interacting with the cell, and every such
intrusion causes a protective stop and a subsequent restart. A small field is
intruded rarely, but, by~\eqref{eq:field}, it admits the same speed only if
$a_{\mathrm{max}}$ is raised, that is, only if the robot is braked more harshly.
For manipulators of large payload the admissible deceleration is bounded by the
drive train and by the permissible load on the fixture and the workpiece, so
that a reduction of the field forces the nominal speed down instead.
Consequently, neither the smallest nor the largest admissible field is
necessarily the most productive one, and the appropriate field depends on the amount of unrelated
traffic that the cell actually experiences. This quantity is not known at
commissioning time, and it is precisely the quantity that the perception layer
is able to observe.

\subsection{Selection of the Safety Configuration}\label{sec:selection}
Let $Z_{1}, \dots, Z_{m}$ denote the candidate configurations of the protective
field that are admissible for the given cell layout, ordered by their radii
$R_{1} < \dots < R_{m}$. For each candidate, the admissible speed cap $v_{j}$ is
obtained by solving~\eqref{eq:field} with equality. Over the observation period, the camera unit records
the number of persons $n(t)$ present in the observed volume together with their
trajectories and, from the recognised activity, whether each of them is engaged
in an activity directed at the cell. The recorded trajectories are then replayed
against each candidate, and the number of protective stops that the
configuration $Z_{j}$ would have caused is counted, separately for persons
working at the cell and for unrelated traffic, which yields an empirical
stop rate $\hat{\lambda}_{j}$. Since all candidates are evaluated on the same
recorded traffic, the comparison is not affected by the variation between
shifts. The recommended configuration follows as:

\begin{equation}\label{eq:select}
j^{*} = \arg\min_{j \in \{1, \dots, m\}}
\left(
c_{\mathrm{s}}\,\hat{\lambda}_{j}
\;+\;
c_{\mathrm{p}}\left(1 - \frac{v_{j}}{v_{\mathrm{nom}}}\right)
\right)
\end{equation}

where $c_{\mathrm{s}}$ is the cost of a protective stop, comprising the stop, the
restart and the reacceleration, and $c_{\mathrm{p}}$ is the cost of operating
below nominal speed, both expressed as lost cycle time. The two terms of~\eqref{eq:select} act
against each other, as the first penalises large fields through the stops they
provoke, whereas the second penalises small fields through the speed cap they
impose. Since $\hat{\lambda}_{j}$ is obtained from the traffic that the cell
actually experiences rather than assumed at commissioning, the recommendation
reflects the use of the shop floor surrounding the cell, which appeared to vary
considerably between the two cells considered here. It should be noted that
$Z_{j^{*}}$ is reported as a recommendation and is not applied automatically.
The occupancy statistics from which it is derived are retained, so that the same
data can additionally serve the data-supported risk assessment for which a clear
demand has been reported \cite{kifz2025mrk}.

\begin{figure*}[tb]
\centering
\includegraphics[width=\textwidth]{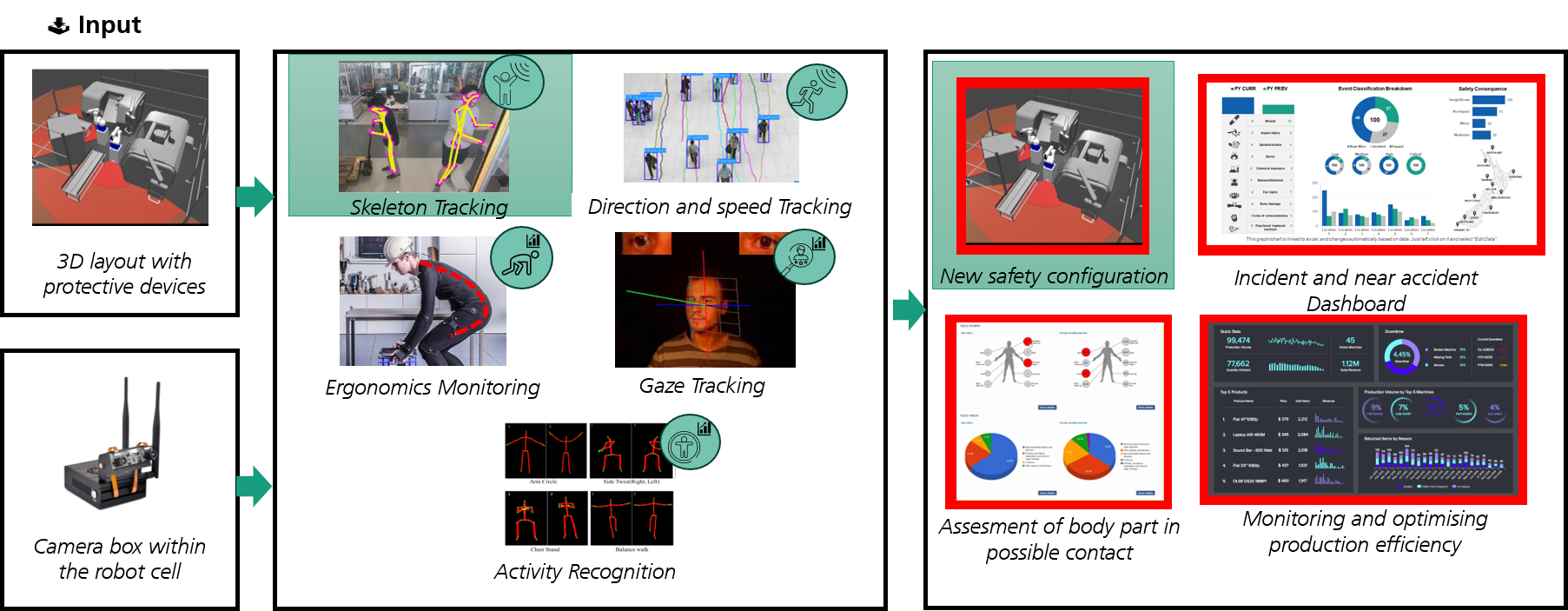}
\caption{The wider concept for the data-supported observation of collaborative
cells. Only the skeleton tracking, the activity recognition and the resulting
proposal for a safety configuration are addressed in this contribution; the
remaining perception modalities and the reporting of key performance indicators
are part of further research.}
\label{fig:outlook}
\end{figure*}

\subsection{Relation to Controller-Side Formulations}
It should be noted that \eqref{eq:ramp} to~\eqref{eq:select} are deliberately elementary. Controller-side
approaches optimise a continuous stop trajectory online from the robot dynamics
and its torque limits, in order to render the safety zone enclosing the
manipulator as small as possible \cite{scalera2022}. In a cell that is already
in production, by contrast, the protective field is physical, and it is fixed at
commissioning, since every configuration of it has to be validated during
acceptance testing. The adaptation therefore cannot take place at the rate of a
supervisory controller. It takes place once, at the level of the configuration
of the cell, and the free variable is not the radius of a zone but the choice
among a small number of admissible configurations. Accordingly, the quantity to
be minimised is not the size of the zone but the expected loss of cycle time
that it causes. The method only needs the stopping distance of the robot at a
given speed, not a full model of the robot dynamics, which is what allows it to
be applied to cells whose controllers are closed.

\subsection{Role of the Perception Unit}
Since the recommendation obtained from~\eqref{eq:select} is reviewed by the safety engineer
and passes through the ordinary acceptance procedure before it becomes
effective, the camera unit does not act upon the cell at any point. An error of the
perception layer therefore degrades the quality of the recommendation, but it
cannot degrade the safety of the cell, and the guarantees of the installed
protective equipment are left intact. This property proved to be a precondition
for the participating companies to permit an installation in an operational
cell.

\section{Results}

The proposed approach was evaluated in several tests, first in a laboratory
cell and subsequently in beta tests conducted together with two industrial
partners. In the beta tests the system was installed in an operational
collaborative cell on the shop floor, so that the perception layer had to
contend with the layouts, occlusions and operator behaviour actually
encountered in production, rather than with a laboratory replica. Owing to
non-disclosure agreements with the participating companies, the underlying
data cannot be published in full; the results are therefore reported in
aggregated and anonymised form.

\subsection{Occupancy Heatmap}
The tracked operator positions accumulated over the observation period are
aggregated into an occupancy heatmap, which represents the relative frequency
with which each region of the cell is occupied by the operator.
Fig.~\ref{fig:heatmap} shows the resulting heatmaps for the two cells, projected
into the cell geometry, where warm colours denote regions of frequent presence
and cold colours denote regions that are rarely or never entered. It can be
observed that the operator presence is concentrated in a compact region in front
of the fixture in both cells, while large parts of the nominal safety zone are
entered only rarely. This spatial imbalance is what the subsequent optimisation
exploits: a protective field dimensioned for the worst case reserves space and
imposes speed reductions for regions in which the operator is in practice seldom
present.

\subsection{Safety Zone Optimisation}
From the occupancy heatmap, an optimal size of the protective separation field
is determined by trading off the achievable robot velocity against the required
safety distance, following the speed and separation monitoring principle of
ISO/TS~15066. Fig.~\ref{fig:optimizer} shows the optimisation for one of the
cells: the left panel overlays the current and the proposed field boundary on
the occupancy heatmap, and the right panel relates the permissible robot
velocity to the safety distance, with the estimated relation used by the
optimiser shown alongside the nominal one. For the cell shown, the protective
field could be reduced from a current safety distance of 1.744~m to 1.435~m
(a reduction of approximately 17\%) over an evaluated cell area of 8.19~m$^2$,
while remaining within the separation requirement. The optimised field is not
applied autonomously to the robot but issued as a proposal that the safety
engineer reviews and, if adopted, incorporates into an updated risk assessment,
so that the responsibility for the safety configuration remains with the
engineer.

\subsection{Productivity}
Over the test period, productivity increased by 9\% on average across the two
instrumented cells.
The magnitude of this effect is considerably smaller than the reductions in
total task time of up to 26.5\% reported for online safety zone scaling under
idealised perception \cite{scalera2022}, which seems plausible given that the
present method reconfigures the protective zone offline from observed occupancy
and leaves the robot controller unmodified, whereas the compared method scales
the zone online from the robot dynamics.

\subsection{Acceptance}
Beyond the measured productivity gain, the beta tests yielded an observation
that is relevant for the practical deployment of such systems. No test customers
beyond the two participating companies could be acquired. Two effects were
observed. Firstly, a considerable effort of persuasion was required in each
acquisition, which was compounded by uncertainty on the side of the companies
regarding data protection, a reservation already reported in
\cite{kifz2025mrk}. Secondly, the participating companies proved reluctant to
modify a safety concept that had already passed acceptance testing, since any
modification entails a renewed risk assessment. This reluctance is not
specific to the present system. In the survey reported in \cite{kifz2025mrk},
only 32\% of the respondents considered a data-supported re-assessment of an
already operating collaborative cell worthwhile, although 66\% expected
safety-relevant data to help them during the risk assessment; the study
attributes this gap to a benefit that has not yet been demonstrated
convincingly in practice and recommends the documentation of deployed
examples, which is what the present contribution provides. The observed
reluctance also indicates where such a method is applied most effectively:
before deployment. If the best-suited configuration is already known at the
planning stage and taken into account during commissioning, the cell enters
operation with a protective field that is optimal with respect to the relevant
performance indicators, and no subsequent modification of an accepted safety
concept is required. It can be concluded that a
demonstrated productivity gain of the observed magnitude does not by itself
constitute a sufficient incentive for adoption, and that a system of this type
has to offer additional value, for instance data-supported risk assessment or
automated documentation, in order to be accepted on the shop floor. It should be
noted, however, that this observation concerns the acceptance of the product and
not the validity of the perception method, which is the subject of this
contribution.

Finally, the evaluation rests on two cells at two companies, and the reported
values should therefore be regarded as indicative rather than statistically
representative.

\section{Conclusion}

This paper presented a camera-based method for supporting the configuration of
collaborative robotic cells. A procedure for deriving the protective field from
the operator occupancy observed in the cell was described, together with the
pose-based perception pipeline from which this occupancy is obtained, and the
method was applied in two operational cells at two industrial partners rather
than only under laboratory conditions. It was shown that the presence of the
operator concentrates in a compact region, while large parts of the nominal
safety zone are entered only rarely, and that the traffic surrounding the cell
differs noticeably between cells of comparable size and equipment, so that a
suitable field configuration does not appear to be readily transferable from
one cell to another. Furthermore, it was argued that the smallest admissible protective
field is not necessarily the most productive one, since a reduction of the field
lowers the frequency of protective stops but raises the deceleration required to
respect the stopping distance, which is particularly restrictive for
manipulators of large payload. The occupancy data required to resolve this
trade-off are not available at commissioning time, and the proposed system
obtains them from the cell in operation. For the cell considered in detail, the
protective field could be reduced by approximately 17\%, and an increase in
productivity of 9\% was measured over the test period.

A larger number of cells, covering a broader range of payloads and layouts, is
required before the trade-off described above can be quantified in general
terms.

\textit{Outlook:}
The method described in this contribution forms part of a wider concept for the
data-supported observation of collaborative cells, which is depicted in
Fig.~\ref{fig:outlook}. Besides the skeleton tracking and the activity
recognition used here, the concept comprises further perception modalities,
namely the tracking of the direction and the speed of the persons approaching
the cell, the monitoring of ergonomic postures, for which the tracked skeleton
could be assessed with RULA \cite{mcatamney1993rula}, and gaze tracking, from
which indicators of the attention of the operator could be derived. These modalities were not
evaluated in the present work. Their combination is expected to yield a richer
description of the interaction than the occupancy statistics used in~\eqref{eq:select}, and it
may permit the recommendation to be conditioned on the recognised phase of the
task rather than on the aggregated traffic alone. Future work should furthermore
examine whether the configuration can be adapted during operation instead of
once at commissioning, which would require the perception layer to be raised to
a safety-rated level and is therefore expected to be considerably more
demanding. Mobile manipulators may offer a natural setting for such an
extension, since they move between workstations whose layout and traffic
differ, so that their protective fields are commonly switched between field sets
during operation anyway, and they already carry cameras for navigation and
manipulation. The reluctance to modify an accepted field configuration observed
in the beta tests would thus largely lose its relevance, and no additional
sensor hardware would be required, so that selecting the field set of each
station from observed occupancy appears to be a promising step once such
platforms are established on the shop floor. Finally, the reservations regarding data protection reported in
\cite{kifz2025mrk} and encountered during the beta tests suggest that on-device
processing without the retention of raw imagery is likely to gain relevance for
the acceptance of such systems.

\bibliographystyle{IEEEtran}
\bibliography{references}

\begin{thebibliography}{10}
\providecommand{\url}[1]{#1}
\csname url@samestyle\endcsname
\providecommand{\newblock}{\relax}
\providecommand{\bibinfo}[2]{#2}
\providecommand{\BIBentrySTDinterwordspacing}{\spaceskip=0pt\relax}
\providecommand{\BIBentryALTinterwordstretchfactor}{4}
\providecommand{\BIBentryALTinterwordspacing}{\spaceskip=\fontdimen2\font plus
\BIBentryALTinterwordstretchfactor\fontdimen3\font minus
  \fontdimen4\font\relax}
\providecommand{\BIBforeignlanguage}[2]{{%
\expandafter\ifx\csname l@#1\endcsname\relax
\typeout{** WARNING: IEEEtran.bst: No hyphenation pattern has been}%
\typeout{** loaded for the language `#1'. Using the pattern for}%
\typeout{** the default language instead.}%
\else
\language=\csname l@#1\endcsname
\fi
#2}}
\providecommand{\BIBdecl}{\relax}
\BIBdecl

\bibitem{villani2018survey}
V.~Villani, F.~Pini, F.~Leali, and C.~Secchi, ``Survey on human-robot
  collaboration in industrial settings: Safety, intuitive interfaces and
  applications,'' \emph{Mechatronics}, vol.~55, pp. 248--266, 2018.

\bibitem{matheson2019hrc}
E.~Matheson, R.~Minto, E.~G.~G. Zampieri, M.~Faccio, and G.~Rosati,
  ``Human-robot collaboration in manufacturing applications: A review,''
  \emph{Robotics}, vol.~8, no.~4, p. 100, 2019.

\bibitem{vysocky2016hrc}
A.~Vysocky and P.~Novak, ``Human-robot collaboration in industry,'' \emph{MM
  Science Journal}, vol. 2016, no.~02, pp. 903--906, 2016.

\bibitem{wang2019symbiotic}
L.~Wang, R.~Gao, J.~V{\'a}ncza, J.~Kr{\"u}ger, X.~V. Wang, S.~Makris, and
  G.~Chryssolouris, ``Symbiotic human-robot collaborative assembly,''
  \emph{CIRP Annals}, vol.~68, no.~2, pp. 701--726, 2019.

\bibitem{gualtieri2021emerging}
L.~Gualtieri, E.~Rauch, and R.~Vidoni, ``Emerging research fields in safety and
  ergonomics in industrial collaborative robotics: A systematic literature
  review,'' \emph{Robotics and Computer-Integrated Manufacturing}, vol.~67, p.
  101998, 2021.

\bibitem{isots15066_2016}
{International Organization for Standardization}, ``{Robots and robotic devices
  --- Collaborative robots},'' ISO, Geneva, Switzerland, Technical
  Specification ISO/TS 15066:2016, 2016.

\bibitem{scalera2022}
L.~Scalera, A.~Giusti, R.~Vidoni, and A.~Gasparetto, ``Enhancing fluency and
  productivity in human-robot collaboration through online scaling of dynamic
  safety zones,'' \emph{The International Journal of Advanced Manufacturing
  Technology}, vol. 121, no.~9, pp. 6783--6798, 2022.

\bibitem{kifz2025mrk}
\BIBentryALTinterwordspacing
T.~Bauernhansl, M.~Huber, and W.~Kraus, ``{Mensch-Roboter-Kollaboration: Status
  quo, Herausforderungen und das Potenzial datenbasierter Risikobeurteilung},''
  KI-Fortschrittszentrum, Fraunhofer IAO und Fraunhofer IPA, Stuttgart,
  Germany, Studie, Studienreihe Lernende Systeme und Kognitive Robotik, 2025,
  accessed 30 June 2026. [Online]. Available:
  \url{https://www.ki-fortschrittszentrum.de/veroeffentlichungen/}
\BIBentrySTDinterwordspacing

\bibitem{zanchettin2016safety}
A.~M. Zanchettin, N.~M. Ceriani, P.~Rocco, H.~Ding, and B.~Matthias, ``Safety
  in human-robot collaborative manufacturing environments: Metrics and
  control,'' \emph{IEEE Transactions on Automation Science and Engineering},
  vol.~13, no.~2, pp. 882--893, 2016.

\bibitem{zanchettin2013path}
A.~M. Zanchettin and P.~Rocco, ``Path-consistent safety in mixed human-robot
  collaborative manufacturing environments,'' in \emph{2013 IEEE/RSJ
  International Conference on Intelligent Robots and Systems (IROS)}, 2013, pp.
  1131--1136.

\bibitem{lasota2015analyzing}
P.~A. Lasota and J.~A. Shah, ``Analyzing the effects of human-aware motion
  planning on close-proximity human-robot collaboration,'' \emph{Human
  Factors}, vol.~57, no.~1, pp. 21--33, 2015.

\bibitem{bdiwi2017new}
M.~Bdiwi, M.~Pfeifer, and A.~Sterzing, ``A new strategy for ensuring human
  safety during various levels of interaction with industrial robots,''
  \emph{CIRP Annals}, vol.~66, no.~1, pp. 453--456, 2017.

\bibitem{hoffman2019evaluating}
G.~Hoffman, ``Evaluating fluency in human-robot collaboration,'' \emph{IEEE
  Transactions on Human-Machine Systems}, vol.~49, no.~3, pp. 209--218, 2019.

\bibitem{iso10218_2_2025}
{International Organization for Standardization}, ``{Robotics --- Safety
  requirements --- Part 2: Industrial robot applications and robot cells},''
  ISO, Geneva, Switzerland, Standard ISO 10218-2:2025, 2025.

\bibitem{iso10218_1_2011}
------, ``{Robots and robotic devices --- Safety requirements for industrial
  robots --- Part 1: Robots},'' ISO, Geneva, Switzerland, Standard ISO
  10218-1:2011, 2011.

\bibitem{marvel2017ssm}
J.~A. Marvel and R.~Norcross, ``Implementing speed and separation monitoring in
  collaborative robot workcells,'' \emph{Robotics and Computer-Integrated
  Manufacturing}, vol.~44, pp. 144--155, 2017.

\bibitem{byner2019dynamicssm}
C.~Byner, B.~Matthias, and H.~Ding, ``Dynamic speed and separation monitoring
  for collaborative robot applications --- concepts and performance,''
  \emph{Robotics and Computer-Integrated Manufacturing}, vol.~58, pp. 239--252,
  2019.

\bibitem{lucci2020combining}
N.~Lucci, B.~Lacevic, A.~M. Zanchettin, and P.~Rocco, ``Combining speed and
  separation monitoring with power and force limiting for safe collaborative
  robotics applications,'' \emph{IEEE Robotics and Automation Letters}, vol.~5,
  no.~4, pp. 6121--6128, 2020.

\bibitem{cao2021openpose}
Z.~Cao, G.~Hidalgo, T.~Simon, S.-E. Wei, and Y.~Sheikh, ``{OpenPose}: Realtime
  multi-person {2D} pose estimation using part affinity fields,'' \emph{IEEE
  Transactions on Pattern Analysis and Machine Intelligence}, vol.~43, no.~1,
  pp. 172--186, 2021.

\bibitem{wojke2017deepsort}
N.~Wojke, A.~Bewley, and D.~Paulus, ``Simple online and realtime tracking with
  a deep association metric,'' in \emph{IEEE International Conference on Image
  Processing (ICIP)}, 2017, pp. 3645--3649.

\bibitem{zhang2022bytetrack}
Y.~Zhang, P.~Sun, Y.~Jiang, D.~Yu, F.~Weng, Z.~Yuan, P.~Luo, W.~Liu, and
  X.~Wang, ``{ByteTrack}: Multi-object tracking by associating every detection
  box,'' in \emph{Computer Vision --- ECCV 2022}, ser. Lecture Notes in
  Computer Science, vol. 13682.\hskip 1em plus 0.5em minus 0.4em\relax
  Springer, 2022, pp. 1--21.

\bibitem{carreira2017i3d}
J.~Carreira and A.~Zisserman, ``Quo vadis, action recognition? {A} new model
  and the {Kinetics} dataset,'' in \emph{IEEE Conference on Computer Vision and
  Pattern Recognition (CVPR)}, 2017, pp. 4724--4733.

\bibitem{feichtenhofer2019slowfast}
C.~Feichtenhofer, H.~Fan, J.~Malik, and K.~He, ``{SlowFast} networks for video
  recognition,'' in \emph{IEEE/CVF International Conference on Computer Vision
  (ICCV)}, 2019, pp. 6201--6210.

\bibitem{yan2018stgcn}
S.~Yan, Y.~Xiong, and D.~Lin, ``Spatial temporal graph convolutional networks
  for skeleton-based action recognition,'' in \emph{AAAI Conference on
  Artificial Intelligence (AAAI)}, vol.~32, 2018.

\bibitem{gu2021harsurvey}
F.~Gu, M.-H. Chung, M.~Chignell, S.~Valaee, B.~Zhou, and X.~Liu, ``A survey on
  deep learning for human activity recognition,'' \emph{ACM Computing Surveys},
  vol.~54, no.~8, pp. 1--34, 2021.

\bibitem{slovak2021vision}
J.~Slov{\'a}k, M.~Melicher, M.~{\v{S}}imovec, and J.~Vach{\'a}lek, ``Vision and
  {RTLS} safety implementation in an experimental human-robot collaboration
  scenario,'' \emph{Sensors}, vol.~21, no.~7, p. 2419, 2021.

\bibitem{bajrami2024multi}
A.~Bajrami, ``Multi {3D}-sensor based human-robot collaboration with cloud
  solution for object handover,'' in \emph{Intelligent Systems and
  Applications. IntelliSys 2023}, ser. Lecture Notes in Networks and Systems,
  K.~Arai, Ed., vol. 825.\hskip 1em plus 0.5em minus 0.4em\relax Cham:
  Springer, 2024, pp. 139--155.

\bibitem{osokin2019lightweight}
D.~Osokin, ``Real-time {2D} multi-person pose estimation on {CPU}: Lightweight
  {OpenPose},'' in \emph{International Conference on Pattern Recognition
  Applications and Methods (ICPRAM)}, 2019, pp. 744--748.

\bibitem{casiez2012oneeuro}
G.~Casiez, N.~Roussel, and D.~Vogel, ``1{\euro} filter: A simple speed-based
  low-pass filter for noisy input in interactive systems,'' in \emph{SIGCHI
  Conference on Human Factors in Computing Systems (CHI)}, 2012, pp.
  2527--2530.

\bibitem{iso13855_2010}
{International Organization for Standardization}, ``{Safety of machinery ---
  Positioning of safeguards with respect to the approach speeds of parts of the
  human body},'' ISO, Geneva, Switzerland, Standard ISO 13855:2010, 2010.

\bibitem{mcatamney1993rula}
L.~McAtamney and E.~N. Corlett, ``{RULA}: A survey method for the investigation
  of work-related upper limb disorders,'' \emph{Applied Ergonomics}, vol.~24,
  no.~2, pp. 91--99, 1993.

\end{thebibliography}

\end{document}